\documentclass[11pt]{article}
\usepackage{acl2015}
\usepackage{times}
\usepackage{url}
\usepackage{latexsym}
\usepackage{comment}
\usepackage{amsmath,amsthm,amssymb}
\usepackage[vlined]{algorithm2e}
\usepackage{graphicx}
\usepackage{multirow}
\usepackage{color}
\usepackage[normalem]{ulem}
\usepackage{skt}
\usepackage[utf8]{inputenc}
\usepackage[russian,english]{babel}
\DeclareMathOperator*{\argmin}{arg\,min}

\newcommand{\vect}[1]{\mathbf{#1}}

\newcommand{\nascomment}[1]{\textcolor{blue}{\bf\small [#1 --NAS]}}

\newcommand{\ignore}[1]{}

\newcommand{\Fref}[1]{Figure~\ref{#1}}

\newcommand{\Tref}[1]{Table~\ref{#1}}

\title{Sparse Overcomplete Word Vector Representations}

\author{Manaal Faruqui  \quad Yulia Tsvetkov \quad Dani Yogatama \quad
  Chris Dyer \quad Noah A. Smith\\
  Language Technologies Institute \\
  Carnegie Mellon University \\
  Pittsburgh, PA, 15213, USA \\
{\tt \{mfaruqui,ytsvetko,dyogatama,cdyer,nasmith\}@cs.cmu.edu}
}

\date{}

\begin{document}
\maketitle

\begin{abstract}
Current distributed representations of words show little
resemblance to theories of lexical semantics.  The former are dense
and uninterpretable, the latter largely based on familiar, discrete
classes (e.g., supersenses) and relations (e.g., synonymy and
hypernymy).  We propose  methods that transform word vectors into
sparse (and optionally binary) vectors.  The resulting representations are more
similar to
the interpretable features typically used in NLP, though they are
discovered automatically from raw corpora.  Because the vectors are
highly sparse, they are computationally easy to work with. Most importantly,
we find that they outperform the original vectors on benchmark tasks.
\end{abstract}

\section{Introduction}

Distributed representations of words have been shown to benefit NLP tasks like
parsing \cite{lazaridou-vecchi-baroni:2013:EMNLP,bansal2014tailoring}, named
entity recognition \cite{guo2014revisiting}, and sentiment
analysis \cite{Socher-etal:2013}. The attraction of word vectors is
that they can be derived directly from raw, unannotated corpora.
Intrinsic evaluations on various tasks are guiding methods toward
discovery of a representation that captures many facts about lexical
semantics \cite{Turney:2001:MWS:645328.650004,Turney10fromfrequency}.

Yet word vectors do not look anything like the representations described in
most lexical semantic theories, which focus on identifying classes of words
\cite{verb-classes.levin.1993,Baker:1998:BFP:980845.980860,Schuler:2005:VBC:1104493} and relationships among word
meanings \cite{miller:1995}.  Though expensive to construct,
conceptualizing word meanings symbolically is important for
theoretical understanding and also when we incorporate lexical
semantics into computational models where interpretability is desired.
On the surface, discrete theories seem
incommensurate with the distributed approach, a problem now receiving
much attention in computational linguistics
\cite{Q13-1015,conf/emnlp/KielaC13,vecchi2013studying,2013arXiv1304.5823G,lewis2014combining,paperno}.

Our contribution to this discussion is a new, principled sparse coding method
that transforms any distributed representation of words into  sparse
vectors, which can then be transformed into binary vectors (\S\ref{sec:framework}). Unlike recent approaches of incorporating semantics
in distributional word vectors
\cite{Yu:2014,xu2014rc,faruqui:2015:retro},  the
method does not rely on any external information source.
The transformation results in longer, sparser vectors, sometimes
called an ``overcomplete'' representation
\cite{Olshausen19973311}. Sparse, overcomplete representations have been
motivated in other domains as a way to increase separability and
interpretability, with each instance (here, a word)
having a small number of active dimensions
\cite{Olshausen19973311,lewicki2000learning}, and to increase stability in the presence of noise \cite{donoho2006stable}.

Our work builds on recent explorations of sparsity as a useful form of
inductive bias in NLP and machine learning more broadly
\cite[\emph{inter
  alia}]{Kazama-Tsujii:2003:EMNLP,N04-1039,friedman2008sparse,glorot2011domain,yogatama2014linguistic}. %\nascomment{pretty
  %sure this is not the Friedman reference you want}
Introducing sparsity in word vector dimensions has been shown to improve
dimension interpretability \cite{nnse,fyshe:2014} and usability of word
vectors as features in downstream tasks \cite{guo2014revisiting}.
The word vectors we
produce are more than 90\% sparse; we also consider binarizing
transformations  that bring them closer to the categories and relations of lexical semantic
theories.
Using a number of state-of-the-art word vectors as input, we find consistent benefits of our method on a suite of
standard benchmark evaluation tasks (\S\ref{sec:expts}).
We also evaluate our word vectors in a word
intrusion experiment with humans
\cite{tealeaves} and find  that our sparse vectors are more interpretable than
the original vectors (\S\ref{sec:intrusion}).

We anticipate that sparse, binary vectors can play an important role
as features in statistical NLP models, which still rely predominantly on
discrete, sparse features whose interpretability enables error
analysis and continued development.   We have made an implementation
of our method publicly available.\footnote{\url{https://github.com/mfaruqui/sparse-coding}}

\section{Sparse Overcomplete Word Vectors}
\label{sec:framework}

\begin{figure*}[!tbp]
  \centering
  \includegraphics[width=1.7\columnwidth]{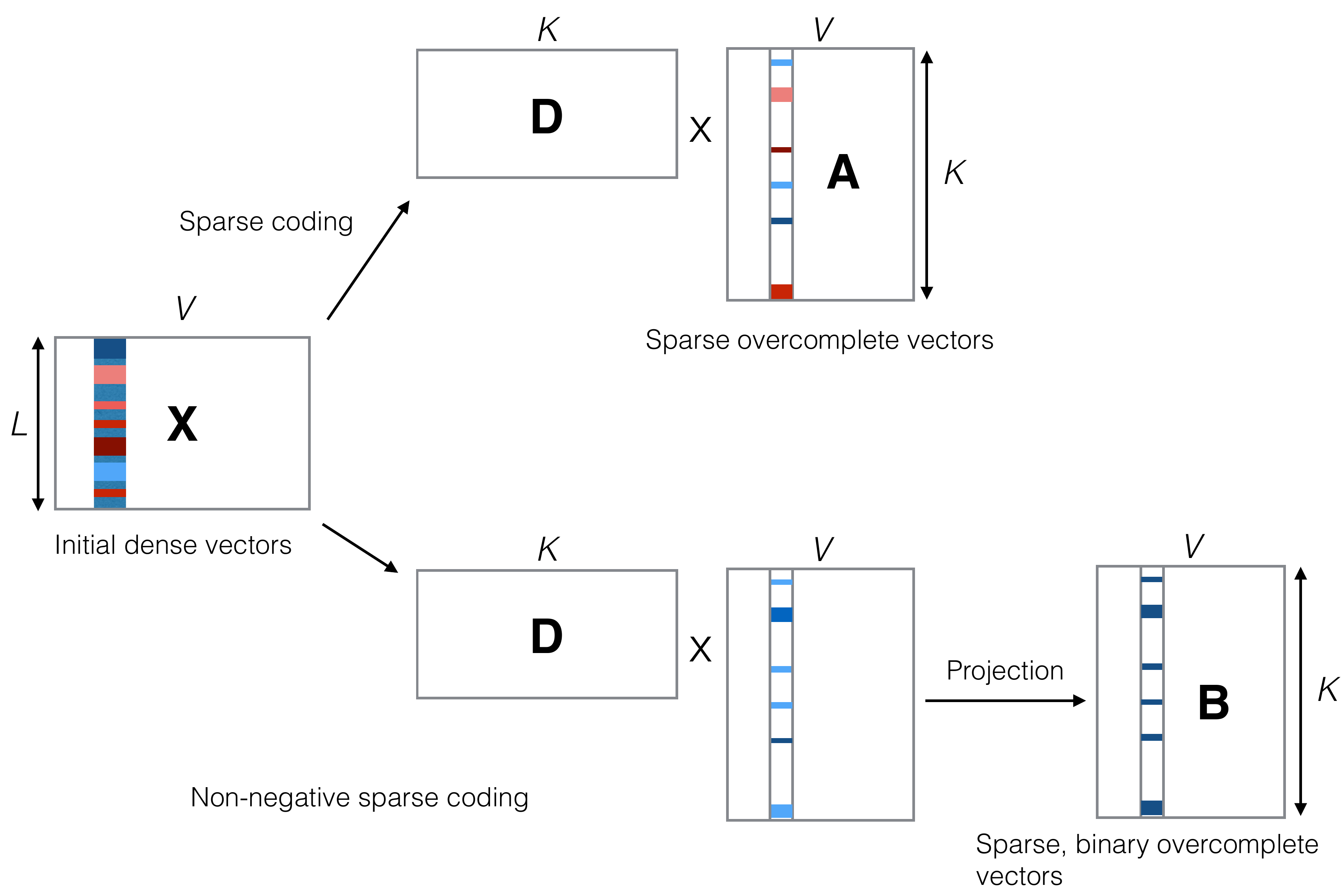}
  \caption{Methods for obtaining sparse overcomplete vectors (top,
    method A, \S\ref{sec:sparse-coding}) and sparse,
  binary overcomplete word vectors (bottom, method B,
  \S\ref{sec:nonneg} and \S\ref{sec:binary}). Observed dense vectors of length
  $L$ (left) are
  converted to sparse non-negative vectors (center) of length $K$ which
  are then projected into the binary vector space (right), where $L \ll K$.
  $\vect{X}$ is dense, $\vect{A}$ is sparse, and $\vect{B}$
  is the binary
  word vector matrix. Strength of colors signify the magnitude of
  values; negative is red, positive is blue, and zero is white.
  \label{fig:schema}}
\end{figure*}

We consider methods for transforming dense word vectors to sparse, binary
overcomplete word vectors. Fig.~\ref{fig:schema} shows two
approaches.  The one on the top, method A,
converts dense vectors to sparse overcomplete vectors
(\S\ref{sec:sparse-coding}).  The one beneath, method B, converts dense vectors to sparse
and binary overcomplete vectors (\S\ref{sec:nonneg} and \S\ref{sec:binary}).
%\nascomment{this figure is confusing.  why are there two paths?  why
% are they not explained?  the lower path seems to match what's in the
% text.  what is the upper path?  why is the font in the figure
% different from the rest of the paper?  why is the source for the
% figure not here and editable? -- reviving this comment.  I don't
% understand why we're confusing the reader with two methods when one
% of them is not ever used in this paper.  I am assuming we will cut
% the ``top'' method in the figure.  I commented out text here that
% discussed the current figure.}

Let $V$ be the vocabulary size.
In the following, $\vect{X}\in
\mathbb{R}^{L \times V}$ is the matrix constructed by stacking $V$
non-sparse ``input'' word vectors of length $L$
(produced by an arbitrary word vector estimator).  We will refer to
these as initializing vectors.
$\vect{A}\in \mathbb{R}^{K \times V}$ contains $V$  sparse overcomplete
word vectors of length $K$.  %\nascomment{not stated before!  please check:}
``Overcomplete'' representation learning implies that $K > L$.

%\cjd{I'm immediately asking: why are you introducing a sparsification procedure in a paper about binary learning? you need to articulate in that this is a tractable relaxation of the binary learning problem, which is NP hard. I'd add a section here giving the ILP formulation and saying rather than trying to solve this, we'll solve an easier problem and then rely on heuristics.}

\subsection{Sparse Coding}
\label{sec:sparse-coding}

In sparse coding~\cite{lee2006efficient}, the goal is to represent each
input vector $\vect{x}_i$ as a sparse linear
combination of basis vectors, $\vect{a}_i$.
Our experiments consider four initializing methods for these vectors, discussed in Appendix~\ref{sec:vectors}.
Given $\vect{X}$, we seek
to solve
\begin{equation}
  \label{equ:loss}
  \argmin_{\vect{D},\vect{A}} \Vert \vect{X} - \vect{D}\vect{A}
  \Vert_2^2 + \lambda \Omega(\vect{A}) + \tau \Vert \vect{D} \Vert_2^2,
\end{equation}
where $\vect{D} \in \mathbb{R}^{L \times K} $ is the dictionary of basis
vectors. %$\mathcal{D}$ is the set of matrices whose columns have small $\ell_2$
%norm \nascomment{this seems redundant since you have the L2
  %regularizer on $\vect{D}$ in the objective function -- I think
  %you can eliminate $\mathcal{D}$ altogether},
$\lambda$ is a regularization hyperparameter, and $\Omega$ is the
regularizer. Here, we use the squared loss for the reconstruction error, but
other loss functions could also be used
\cite{honglak}. %\nascomment{cut these if we've already said it
  %before, as I propose above:  Note that it is not
%necessary, although typical, for $K$ to be less than $L$. In case of
%overcomplete representations: $K > L$, which is the subject of our paper.}
To obtain sparse word representations we will impose an $\ell_1$ penalty on
$\vect{A}$. Eq.~\ref{equ:loss} can be broken down into loss for each word
vector which can be optimized separately in parallel (\S\ref{sec:sparse-inference}):
\begin{equation}
  \label{equ:sparse}
  \argmin_{\vect{D},\vect{A}} \sum_{i=1}^{V}
  \Vert \vect{x}_{i} - \vect{D}\vect{a}_i
  \Vert_{2}^{2} + \lambda \Vert \vect{a}_i \Vert_1 + \tau \Vert \vect{D}
  \Vert_2^2
\end{equation}
where $\vect{m}_i$ denotes the $i$th column vector
of matrix $\vect{M}$. %\nascomment{check; I made this more abstract
  %since we use the same notation for $\vect{A}$ and $\vect{D}$.}
%\mfar{What in $\vect{M}$, $\vect{m}_i$?}
Note that this problem is not convex.  We refer to this approach as
\textbf{method A}.

\subsection{Sparse Nonnegative Vectors}
\label{sec:nonneg}

Nonnegativity in the feature space has often been shown to correspond to
interpretability
\cite{lee1999learning,cichocki2009nonnegative,nnse,fyshe:2014,fyshe:2015}.
To obtain nonnegative sparse word vectors, we use a variation of the
nonnegative sparse coding method \cite{hoyer2002non}.
Nonnegative sparse coding further constrains the problem in
Eq.~\ref{equ:sparse} so that
$\vect{D}$ and $\vect{a}_i$ are nonnegative.  Here, we
apply this constraint only to the representation vectors
$\{\vect{a}_i\}$.
%\nascomment{I don't know why is this cited here:} \cite{nnse}.
Thus, the new objective for nonnegative sparse vectors becomes:
\begin{equation}
  \label{equ:nonneg}
 \argmin_{\vect{D} \in \mathbb{R}_{\geq 0}^{L \times K},\vect{A} \in \mathbb{R}_{\geq 0}^{K \times V}}
  \sum_{i=1}^{V}\Vert \vect{x}_{i} - \vect{D}\vect{a}_i \Vert_{2}^{2}
  + \lambda \Vert \vect{a}_i \Vert_1 + \tau \Vert \vect{D}
  \Vert_2^2
\end{equation}
This problem will play a role in our second
approach, \textbf{method B}, to which we will return shortly.
This nonnegativity constraint can be easily incorporated during optimization,
as explained next.
%We further employ another constraint on the objective to obtain a novel model of sparse word vector representations containing only binary 0, 1 elements i.e, $\vect{a}_i \in \mathbb{R}_{\{0,1\}}^{K}$. These binary vectors can easily be obtained by converting all positive elements of the nonnegative vectors to 1 and retaining all the zero elements as is.
%Binary vectors occupy much less memory as they can be represented as
%series of on/off bits and are also much faster in vector computations
%as later explained in \S\ref{sec:analysis}.

\subsection{Optimization}
\label{sec:sparse-inference}

We use online adaptive gradient descent (AdaGrad;
\nocite{Duchi:EECS-2010-24} Duchi et al., 2010)
for solving the optimization problems in Eqs.~\ref{equ:sparse}--\ref{equ:nonneg} by updating $\vect{A}$ and $\vect{D}$.
%\nascomment{giveformula numbers, I'm not sure which two you are talking about}.
\ignore{
AdaGrad provides a per-feature \nascomment{what is a
  ``feature''?  you are using terminology from some other optimization
  problem that AdaGrad is applied to, and it is very confusing.
  explain what variables are being optimized here.  D? A? both?} learning rate at each time
step $t$,
\begin{equation*}
  \eta_t = \frac{\eta}{\sqrt{\sum_{t' = 1}^{t} g_{t'}^{2}}}
\end{equation*}
where $g_{t'}$ is the gradient at step $t'$.
 This is used for updating
$\vect{D}$ as the objective is convex and differentiable in $\vect{D}$.}
In order to speed up training
we use asynchronous updates to the parameters of the model in parallel
for every word vector \cite{duchi:dual,42248}.

However, directly applying stochastic subgradient descent to an
$\ell_1$-regularized objective fails to produce sparse solutions
in bounded time, which has motivated several specialized algorithms that target
such objectives. We use the AdaGrad variant of one such learning
algorithm, the regularized dual averaging
algorithm \cite{Xiao09dualaveraging}, which keeps track of the online average
gradient at time $t$:
%\begin{equation*}
  $\bar{g}_t = \frac{1}{t} \sum_{t'=1}^{t} g_{t'}$
%\end{equation*}
Here, the subgradients do not include terms for the regularizer; they are
derivatives of the unregularized objective ($\lambda = 0, \tau = 0$)
%\nascomment{confusing; setting lambda = 0 does not get rid of the
%  other regularization term, for D.  this is very confusing})
with respect to $\vect{a}_i$. We define
\begin{equation*}
\gamma = -\textrm{sign}(\bar{g}_{t, i, j}) \frac{\eta t}{\sqrt{G_{t,i,j}}}
    (|\bar{g}_{t, i, j}|-\lambda),
\end{equation*}
where  $G_{t,i,j} = \sum_{t'= 1}^{t} g_{t',i,j}^{2}$.
Now, using the average gradient, the $\ell_1$-regularized
objective is optimized as follows:
\begin{equation}
  a_{t+1,i,j} =
  \begin{cases}
    0, & \mbox{if $|\bar{g}_{t, i, j}| \leq \lambda$}\\
    \gamma, &\text{otherwise}\\
  \end{cases}
\end{equation}
where, $a_{t+1,i,j}$ is the $j$th element of sparse %\nascomment{what
  %if it's not sparse at this moment?  don't call it ``sparse''}
 vector $\vect{a}_i$
at the $t$th update and $\bar{g}_{t, i, j}$ is the corresponding average
gradient.
For obtaining nonnegative sparse vectors we take projection of the updated
$\vect{a}_i$ onto $\mathbb{R}_{\geq 0}^{K}$ by choosing the closest point in
$\mathbb{R}_{\geq 0}^{K}$ according to Euclidean distance (which corresponds
to zeroing out the negative elements):
\begin{equation}
  a_{t+1,i,j} =
  \begin{cases}
    0, &\mbox{if $|\bar{g}_{t, i, j}| \leq \lambda$}\\
    0, & \mbox{if $\gamma < 0$}\\
    \gamma, &\mbox{otherwise}
  \end{cases}
\end{equation}
%This can be seen as an application of the ReLU non-linearity during
%optimization \cite{glorot2011deep}. \nascomment{I don't understand
%this point.  why are we saying that it ``can be seen'' this way?

\begin{table}[!tb]
  \centering
  \begin{tabular}{|lr|rrr|r|}
  \hline
  $\vect{X}$ & $L$ &
$\lambda$ & $\tau$  & $K$ & \% Sparse\\% & \multicolumn{2}{|c|}{WS-353}\\
  \hline
  Glove & $300$ & $1.0$ & $10^{-5}$ & $3000$ & $91$ \\% & $60.5$ & $62.8$\\
  SG & $300$ & $0.5$ & $10^{-5}$  & $3000$ & $92$ \\%& $65.5$ & $59.6$\\
  GC & $50$ & $1.0$ & $10^{-5}$ & $500$ & $98$ \\%& $62.3$ & $64.4$\\
  Multi & $48$ & $0.1$ & $10^{-5}$ & $960$ & $93$ \\% & $58.1$ & $58.7$\\
  \hline
  \end{tabular}
\caption{Hyperparameters for learning sparse overcomplete vectors tuned on
  the WS-353 task.  Tasks are explained in \S\ref{sec:benchmarks}.  The four initial
  vector representations  $\vect{X}$ are explained in
  \S\ref{sec:vectors}. \ignore{\nascomment{the
    last two columns are not clearly explained; I think they are
    correlation scores on the task.  if so, why only ``dense'' and
    binary?  why not sparse as well?  which one was the tuning criterion?  whatever they are, the caption
    needs to make that clear.}}}
  \label{tab:hyper}
\end{table}

\subsection{Binarizing Transformation}
\label{sec:binary}

\begin{table}[!tb]
  \centering
  \begin{tabular}{|l|}
%  \hline
 % Cluster ID & Words\\
  \hline
%  67 &
 hot, fresh, fish, 1/2, wine, salt\\ \hline
 % 22 &
series, tv, appearances, episodes\\ \hline
%  14 &
1975, 1976, 1968, 1970, 1977, 1969\\ \hline
%  45 &
 dress, shirt, ivory, shirts, pants \\ \hline
%  1 &
upscale, affluent, catering, clientele\\
  \hline
  \end{tabular}
    \caption{Highest frequency words in randomly picked word clusters of
    binary sparse overcomplete Glove vectors.
  \label{tab:clusters}}
\end{table}

Our aim with \textbf{method B}  is to obtain word representations that can emulate
the binary-feature space designed for various NLP tasks. We could
state this as an optimization problem:
\begin{equation}
  \label{equ:ilp}
 \argmin_{\stackrel{\vect{D} \in \mathbb{R}^{L \times K}}{\vect{B} \in
     \{0,1\}^{K \times V}}}
  \sum_{i=1}^{V}\Vert \vect{x}_{i} - \vect{D}\vect{b}_i \Vert_{2}^{2}
  + \lambda \Vert \vect{b}_i \Vert_1^1 + \tau \Vert \vect{D} \Vert_2^2
\end{equation}
where $\vect{B}$ denotes the binary (and also sparse) representation.
This is an mixed integer bilinear program, which is NP-hard
\cite{alkhayyal:1983}.
Unfortunately, the number of variables in the problem
is $ \approx KV$ which reaches $100$ million when $V = 100,000$
and $K = 1,000$, which is intractable to solve using standard techniques.

A more tractable relaxation to this hard
 problem is to first constrain the continuous
representation $\vect{A}$ to be
nonnegative (i.e, $\vect{a}_i \in \mathbb{R}_{\geq 0}^{K}$;
\S\ref{sec:nonneg}). Then, in order to avoid an expensive computation, we take the nonnegative word
vectors obtained using Eq.~\ref{equ:nonneg} and project
nonzero values to $1$, preserving the $0$ values.  Table~\ref{tab:clusters} shows a random set of word
clusters obtained by (i) applying our method to Glove initial vectors
and (ii) applying $k$-means clustering ($k = 100$).
 In
\S\ref{sec:expts} we will find that these vectors perform well
quantitatively.

\subsection{Hyperparameter Tuning}
\label{sec:tune}
%\nascomment{I moved this.  ok?}
Methods A and B have three hyperparameters:
the $\ell_1$-regularization penalty
$\lambda$, the $\ell_2$-regularization penalty $\tau$, and the length
of the overcomplete word vector representation $K$.
We perform a grid search on $\lambda \in \{0.1, 0.5, 1.0\}$ and
$K \in \{10L, 20L\}$, selecting values that maximizes performance
on one ``development'' word similarity task (WS-353, discussed in
\S\ref{sec:benchmarks}) while achieving at least 90\% sparsity in
overcomplete vectors.
$\tau$ was tuned on one collection of initializing vectors  (Glove, discussed in
\S\ref{sec:vectors}) so that the
vectors in $\vect{D}$ are near unit norm. %\nascomment{for what criterion?}
The four vector representations and their corresponding hyperparameters selected
by this procedure are summarized in Table~\ref{tab:hyper}. These hyperparameters
were chosen for method A and retained for method B.
%\nascomment{were these chosen using method 1 or method 2?}

%\nascomment{this does not belong here:  An added advantage of sparse binary word vectors is that they can be stored as bit-vectors instead of storing them as real numbers which leads to smaller memory footprint.}

%For skip-gram
%vectors, we could not obtain an improvement in performance on WS-353 with
%binary vectors and hence we tuned the parameters to minimize the loss in
%performance.
%\nascomment{somewhere around here it needs to be discussed that
%  skipgram sparse binary (?) vectors are worse than the original ones}

\section{Experiments}
\label{sec:expts}

%\nascomment{Again, as written right now, I can't tell if you're using
%  the nonnegative sparse method here or not.  the first figure makes
%  it totally unclear}

Using methods A and B, we constructed sparse overcomplete vector
representations $\vect{A}$ and $\vect{B}$ resp.,
starting from four initial vector representations
$\vect{X}$; these are explained in Appendix~\ref{sec:vectors}.  We
used one benchmark evaluation (WS-353) to tune hyperparameters,
resulting in the settings shown in Table~\ref{tab:hyper}; seven other
tasks were used to evaluate the quality of the sparse overcomplete
representations.  The first of these is a word similarity task, where
the score is correlation with human judgments, and the others are
classification accuracies of an $\ell_2$-regularized logistic
regression model trained using the word vectors.  These tasks are described
in detail in Appendix~\ref{sec:benchmarks}.

\begin{table*}[!tbh]
  \centering
  \begin{tabular}{|ll|r|rrrrr|r|r|}
  \hline
  \multicolumn{2}{|c|}{\multirow{2}{*}{Vectors}}& SimLex & Senti.& TREC & Sports & Comp. & Relig. & NP & \multirow{2}{*}{Average}\\
& & Corr.  & Acc. & Acc. & Acc. & Acc. & Acc. & Acc. &\\
\hline
\multirow{3}{*}{Glove} & $\vect{X}$ & $36.9$ & $77.7$ & $76.2$ & $95.9$ & $79.7$ & $86.7$ & $77.9$ & $76.2$\\
%\hline
& $\vect{A}$ & $38.9$  & $\textbf{81.4}$ & $\textbf{81.5}$ & $\textbf{96.3}$ & $\textbf{87.0}$ & $\textbf{88.8}$ & $\textbf{82.3}$ & $\textbf{79.4}$\\
%NNOC & $3000$ & $60.5$ & $39.5$ & $73.0$ & $81.2$ & $81.4$ & $\textbf{81.8}$\\
& $\vect{B}$ & $\textbf{39.7}$  & $81.0$ & $81.2$ & $95.7$ & $84.6$ & $87.4$ & $81.6$ & $78.7$\\
%\multicolumn{2}{|l}{Glove + \newcite{guo2014revisiting}} & $300$ & $58$ & $39.4$ & $78.2$ & $77.8$ & $95.9$ & $77.2$ & $85.0$ & $77.5$\\
%\multicolumn{2}{|l}{\newcite{nnse}} & $300$ & $86$  & $27.1$ & $73.7$ & $74.0$ & $95.9$ & $77.4$ & $82.8$ & $75.5$\\
\hline
\multirow{3}{*}{SG} & $\vect{X}$ & $\textbf{43.6}$ & $81.5$ & $77.8$ & $97.1$ & $80.2$ & $85.9$ & $80.1$ & $78.0$\\
%\hline
& $\vect{A}$& $41.7$  & $\textbf{82.7}$ & $81.2$ & $\textbf{98.2}$ & $84.5$ & $86.5$ & $81.6$ & $79.4$\\
%NNOC & $3000$ & $57.7$ & $43.6$ & $65.5$ & $\textbf{82.7}$ & $81.4$ &
%$\textbf{82.0}$\\
& $\vect{B}$ & $42.8$ & $81.6$ & $\textbf{81.6}$ & $95.2$ & $\textbf{86.5}$ & $\textbf{88.0}$ & $\textbf{82.9}$ & $\textbf{79.8}$\\
\hline
\multirow{3}{*}{GC} & $\vect{X}$ & $9.7$  & $68.3$ & $64.6$ & $75.1$ & $60.5$ & $76.0$ & $79.4$ & $61.9$\\
%\hline
& $\vect{A}$ & $12.0$ & $73.3$ & $77.6$ & $77.0$ & $68.3$ &
$\textbf{81.0}$ & $\textbf{81.2}$ & $67.2$\\
%NNOC & $500$ & $\textbf{65.1}$ & $\textbf{20.9}$ & $\textbf{39.1}$ & $72.8$ &
%$\textbf{81.0}$ & $\textbf{82.1}$\\
& $\vect{B}$ & $\textbf{18.7}$ & $\textbf{73.6}$ & $\textbf{79.2}$
& $\textbf{79.7}$ & $\textbf{70.5}$ & $79.6$ & $79.4$ & $\textbf{68.6}$\\
\hline
\multirow{3}{*}{Multi} & $\vect{X}$ & $\textbf{28.7}$  & $75.5$ & $63.8$ & $83.6$ & $64.3$ & $81.8$ & $79.2$ & $68.1$\\
%\hline
& $\vect{A}$ & $28.1$ & $\textbf{78.6}$ & $79.2$ & $93.9$ & $78.2$ &
$84.5$ & $81.1$ & $74.8$\\
%NNOC & $960$ & $58.3$ & $26.7$ & $65.2$ & $78.5$ & $79.0$ & $81.9$\\
& $\vect{B}$ &$\textbf{28.7}$  & $77.6$ & $\textbf{82.0}$ &
$\textbf{94.7}$ & $\textbf{81.4}$ & $\textbf{85.6}$ & $\textbf{81.9}$ & $\textbf{75.9}$\\
\hline
\end{tabular}
\caption{Performance comparison of transformed vectors to initial
  vectors $\vect{X}$.  We show sparse overcomplete representations
  $\vect{A}$ and also
  binarized representations $\vect{B}$.  Initial vectors are discussed in
  \S\ref{sec:vectors} and tasks in
  \S\ref{sec:benchmarks}. %\nascomment{why are correlations greater
    %than 1?  are you using a 100 point scale? why?  need to clarify.}
\label{tab:results}}
\end{table*}

\subsection{Effects of Transforming Vectors}

First, we quantify the effects of our transformations by comparing
their output to the initial ($\vect{X}$) vectors.
Table~\ref{tab:results} shows consistent improvements of sparsifying
vectors (method A).  The exceptions are on the SimLex task, where our sparse
vectors are worse than the skip-gram initializer and on par with the
multilingual initializer.   Sparsification is beneficial across all of
the text classification tasks, for all initial vector
representations.  On average across all vector types and all tasks,
sparse overcomplete vectors outperform their corresponding
initializers by 4.2 points.\footnote{We report correlation on a 100 point
scale, so that the average which includes accuracuies and correlation
is equally representatitve of both.}% \nascomment{check}}

Binarized vectors (from method B) are also usually better than the initial vectors
(also shown in Table~\ref{tab:results}), and tend to outperform the
sparsified variants, except when initializing with Glove.
On average across all vector types and all tasks,
binarized overcomplete vectors outperform their corresponding
initializers by 4.8 points and the continuous, sparse
intermediate vectors by 0.6 points.

%\nascomment{From here on, I'm a little confused -- but I THINK we are
  %only using sparse, not binarized vectors.  we should say that:}
  From here on, we explore more deeply the sparse overcomplete
  vectors from method A (denoted by $\vect{A}$), leaving binarization
  (method B) aside.

\subsection{Effect of Vector Length}

\begin{figure}[!tb]
  \centering
  \includegraphics[width=.9\columnwidth]{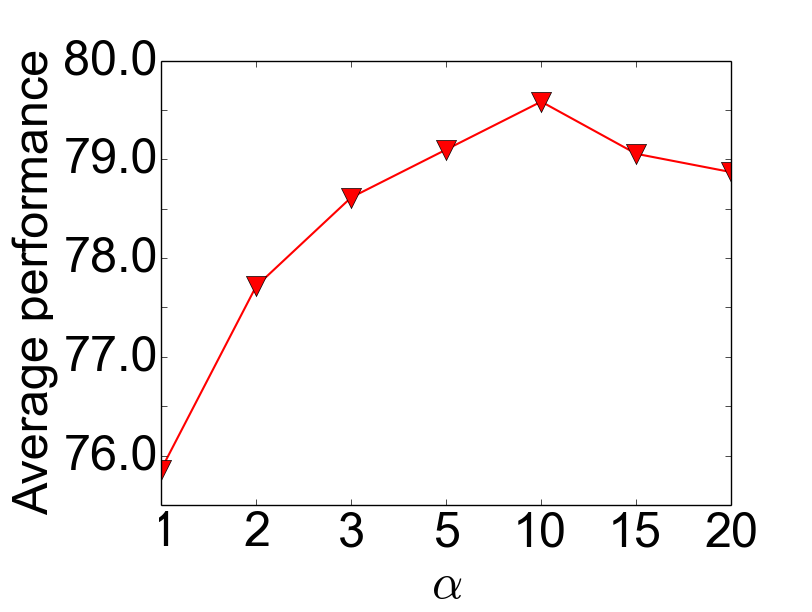}
  \caption{Average performace across all tasks for
  %\nascomment{binarized?  not clear}
sparse overcomplete vectors ($\vect{A}$) produced by
    Glove initial vectors, as a function of the ratio of $K$ to $L$.}
    %\nascomment{rename the x-axis to $\alpha$ to match text.
    %  ``sparsification factor'' is misleading; it is not the extent to
    %which the vectors are ``sparsified'' but rather has to do with the length of the new vector}
  \label{fig:length}
\end{figure}

How does the length of the overcomplete vector ($K$) affect
performance?  We focus here on the Glove vectors, where $L=300$, and report
average performance across all tasks.  We
consider $K = \alpha L$ where $\alpha \in \{1, 2, 3, 5, 10, 15, 20\}$.
\Fref{fig:length} plots the average performance across tasks against
$\alpha$.  The earlier selection of $K = 3,000$ ($\alpha = 10$) gives
the best result; gains are monotonic in $\alpha$ to that point and
then begin to diminish.

\subsection{Alternative Transformations}

We consider two alternative transformations.  The first preserves the
original vector length but achieves a binary, sparse vector ($\vect{B}$) by
applying:
\begin{equation} \label{eq:thresh1}
b_{i,j} = \left\{ \begin{array}{ll} 1 & \mbox{if $x_{i,j} > 0$} \\ 0 &
    \mbox{otherwise} \end{array} \right.\end{equation}
%\nascomment{I used $b$ and $x$ above but perhaps it should be $a, x$
%  or $b, a$.  Not clear what your inputs and outputs are!}

The second transformation was proposed by \newcite{guo2014revisiting}.  Here, the original vector
length is also preserved, but sparsity is achieved through:
\begin{equation} \label{eq:thresh2}
a_{i,j} = \left\{ \begin{array}{rl} 1 & \mbox{if $x_{i,j} \ge M^+$} \\
      -1 & \mbox{if $x_{i,j} \le M^-$} \\
0 &
    \mbox{otherwise} \end{array} \right. \end{equation}
where $M^+$ ($M^-$) is the mean of positive-valued (negative-valued) elements of
$\vect{X}$.  These vectors are, obviously, not binary.

We find that on average, across initializing vectors and across all tasks %\nascomment{all
  %or just the text classification tasks?  best to be consistent with
  %the earlier parts of the paper and explain why when we are not
  %consistent},
that our sparse overcomplete ($\vect{A}$)
%\nascomment{and binary? not clear!}
vectors lead to better performance than either of the alternative
transformations.

\begin{table*}[!tb]
  \centering
  \begin{tabular}{|l|rrrr|r|}
  \hline
  \multicolumn{1}{|r|}{$\vect{X}$:} & Glove & SG & GC & Multi & Average\\
  \hline
  $\vect{X}$ & $76.2$ & $78.0$ & $61.9$& $68.1$ & $71.0$\\
  \hline
  Eq.~\ref{eq:thresh1} & $75.7$ & $75.8$ & $60.5$ & $64.1$ & $69.0$\\
  Eq.~\ref{eq:thresh2} \cite{guo2014revisiting} & $75.8$ & $76.9$ & $60.5$ &  $66.2$ & $69.8$\\
\hline
  $\vect{A}$ & $\textbf{79.4}$ & $\textbf{79.4}$ & \textbf{67.2} & \textbf{74.8} & \textbf{75.2}\\
  \hline
  \end{tabular}
    \caption{Average performance across all tasks and vector
      models using different transformations. %\nascomment{not clear
        %which of ``our methods'' this is.  I assumed sparse but not binarized}
  \label{tab:compare}}
\end{table*}

\section{Interpretability}
\label{sec:intrusion}

%\nascomment{these two things (before and after ``i.e.'' are not the
  %same.  we can't say anything at all about ``certain semantic or
  %syntactic properties of a word'' from these experiments.  cut that.}
Our hypothesis is that the dimensions of sparse overcomplete vectors are more
interpretable than those of dense word vectors.\ignore{ i.e, the dimensions
in the sparse word vectors correspond to certain semantic or syntactic
properties of a word.}
Following \newcite{nnse}, we use a word intrusion experiment \cite{tealeaves}
to corroborate this hypothesis. In addition, we conduct qualitative analysis
of interpretability, focusing on individual dimensions.

\subsection{Word Intrusion}
%\nascomment{I don't like how this is explained.  first tell us what
%  you're trying to evaluate, intuitively.  then tell us the
%  procedure.  this is very hand-wavey; I could not replicate your
%  experiment from this level of detail.}
Word intrusion experiments seek to quantify the extent to which
dimensions of a learned word representation are coherent to humans.
In one instance of the experiment, a human judge is
presented with five words in random order and asked to select the
``intruder.'' The words are selected by the experimenter by choosing
%\nascomment{uniformly at random?}
one dimension $j$ of the learned representation, then ranking the
words on that dimension alone. The dimensions are chosen in decreasing order
of the variance of their values across the vocabulary.
Four of the words are the top-ranked
words according to $j$, and the ``true'' intruder is a word from the
bottom half of the list, chosen to be a word that appears in the top
10\% of some other dimension.  An example of an instance is:
\begin{center}
\em naval, industrial, technological, marine, identity
\end{center}
(The last word is the intruder.)

We formed instances from initializing vectors and from our sparse
overcomplete vectors ($\vect{A}$). Each of these two
combines the four different initializers $\vect{X}$.  We selected
the 25 dimensions $d$ in each case.
Each of the 100  instances per condition (initial vs.~sparse overcomplete) was
given to three judges. %\nascomment{the way this is
  %written, readers will assume A1 is the same person throughout, and
  %he looked at every instance.  same with A2 and A3.  if that is not
  %correct, then the table and the discussion need more work.}

\begin{table}[!tb]
  \centering
  \begin{tabular}{|l|rrr|r|rr|}
  \hline
  Vectors & A1 & A2 & A3 & Avg. & IAA & $\kappa$\\
  \hline
  $\vect{X}$ & $61$ & $53$ & $56$ & $57$ & $70$ & $0.40$\\
  $\vect{A}$ & $71$ & $70$ & $72$ & $71$ & $77$ & $0.45$\\
  \hline
  \end{tabular}
    \caption{Accuracy of three human annotators on the word intrusion task,
             along with the average inter-annotator agreement \cite{artstein2008inter}
             and Fleiss' $\kappa$ \cite{davies1982measuring}.
  \label{tab:intrusion}}
\end{table}

Results in \Tref{tab:intrusion} confirm that the sparse overcomplete vectors are more
interpretable than the dense vectors. The inter-annotator agreement
on the sparse vectors increases substantially, from $57\%$ to $71\%$,
and the Fleiss' $\kappa$ increases from ``fair'' to ``moderate''
agreement \cite{landis1977measurement}.

\subsection{Qualitative Evaluation of Interpretability}
If a vector dimension is interpretable, the top-ranking words for that dimension
should display semantic or syntactic groupings. To verify this qualitatively,
we select five dimensions with the
highest variance of values in initial and sparsified GC vectors.
We compare top-ranked words in the dimensions extracted from the
two representations. The words are listed in \Tref{tab:words}, a dimension
per row.  Subjectively, we find the semantic groupings better in the sparse
vectors than in the initial vectors.
\begin{table}[htb]
  \centering
  \begin{tabular}{|l|l|}
  %\hline
  %Vectors & Words\\
  \hline
  \multirow{5}{*}{$\vect{X}$} & combat, guard, honor, bow, trim, naval\\
  & 'll, could, faced, lacking, seriously, scored\\
  & see, n't, recommended, depending, part\\
  & due, positive, equal, focus, respect, better \\
  & sergeant, comments, critics, she, videos \\
  \hline
  \multirow{5}{*}{$\vect{A}$} & fracture, breathing, wound, tissue, relief\\
  & relationships, connections, identity, relations\\
  & files, bills, titles, collections, poems, songs\\
  & naval, industrial, technological, marine \\ %, physics\\
  & stadium, belt, championship, toll, ride, coach\\
  \hline
  \end{tabular}
    \caption{Top-ranked words per dimension for initial and  sparsified
    GC representations. Each line shows words from a different dimension.}
  \label{tab:words}
\end{table}

\begin{figure*}[!tb]
  \centering
  \includegraphics[scale=0.45]{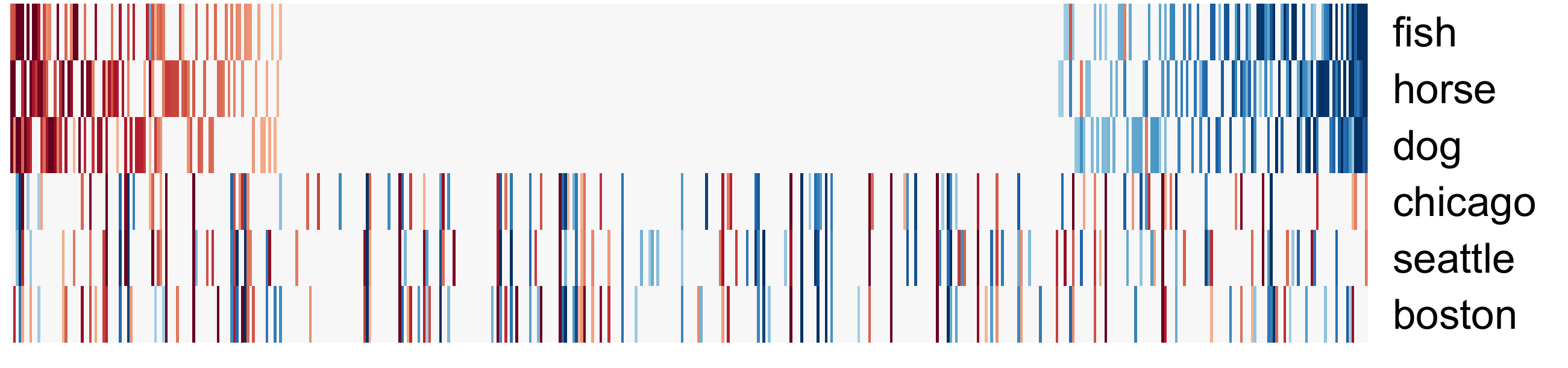}
  \caption{Visualization of sparsified GC vectors.  Negative values are red, positive values are
  blue, zeroes are white.}
  %Strength of colors signify the magnitude of values, with
  %red as negative, blue as positive sign and white as zero.}
  \label{fig:heatmap}
\end{figure*}

Figure~\ref{fig:heatmap} visualizes the sparsified GC vectors for six
words.  The dimensions are sorted by the average value across the
three ``animal'' vectors.  The animal-related words use many of the
same dimensions (102 common active dimensions out of 500 total);
in constrast, the three city names use mostly
distinct vectors. %\nascomment{quantify this; how much overlap?}

% To further understand whether overcomplete representations
% are interpretable, we visualize them
% with a heat map of vector values. As in the previous experiments,
% we expect words within a semantic grouping to have a common salient
% dimension with high values.
% We select a semantic grouping \textsc{animal}
% with GC-sparse word vectors \textit{fish}, \textit{horse} and \textit{dog}.
% To construct a prototypical \textsc{animal} vector, we average the vectors.
% Then, we sort the dimensions in the prototypical vector by magnitude.
% \Fref{fig:heatmap} plots the vectors in the obtained dimension ordering.
% We contrast these with the similarly sorted GC-sparse word vectors from another
% semantic grouping: \textsc{location} with vectors \textit{chicago}, \textit{seattle}
% and \textit{boston}. A contrasting pattern in the vector dimensions of these two
% categories is evident.

\section{Related Work}
\label{sec:related}

To the best of our knowledge, there has been no prior work on obtaining
overcomplete word vector representations that are sparse and
categorical. However, overcomplete features have been widely used in image
processing, computer vision \cite{Olshausen19973311,lewicki2000learning}
and signal processing \cite{donoho2006stable}. Nonnegative matrix
factorization is often used for interpretable coding of information
\cite{lee1999learning,liu2003non,cichocki2009nonnegative}.

Sparsity constraints are in general useful in NLP problems
\cite{Kazama-Tsujii:2003:EMNLP,friedman2008sparse,N04-1039},
like POS tagging \cite{ganchev2009}, dependency parsing
\cite{Martins:2011}, text classification \cite{yogatama2014linguistic}, and
representation learning \cite{bengio2013representation,yogatama:2015}.
Including sparsity constraints in Bayesian models of lexical semantics
like LDA in the form of sparse Dirichlet priors has been shown to be
useful for downstream tasks like POS-tagging \cite{toutanova2007bayesian}, and
improving interpretation \cite{paul2012factorial,zhu2012sparse}.

\section{Conclusion}
\label{sec:conclusion}

We have presented a method that converts word vectors
obtained using any state-of-the-art word vector model into
sparse and optionally binary word vectors.
These transformed vectors  appear to come closer to features
used in NLP tasks and outperform the original  vectors from which they
are derived on a suite of
semantics and syntactic evaluation benchmarks. We also find that the sparse
vectors are more interpretable than the dense vectors by humans according to a
word intrusion detection test.

\section*{Acknowledgments}

We thank Alona Fyshe for discussions on vector interpretability
and three anonymous reviewers for their feedback.
This research was supported in part by the National Science Foundation through
grant IIS-1251131 and  the Defense
Advanced Research Projects Agency through grant
FA87501420244.
This work was supported in part by the U.S. Army Research Laboratory and the U.S. Army Research Office under contract/grant number W911NF-10-1-0533.

\appendix
\section{Initial Vector Representations ($\vect{X}$)} \label{sec:vectors}

Our experiments consider four publicly available collections of
pre-trained word vectors.
They vary in the amount of data used and the estimation method.

\paragraph{Glove.} Global vectors for word representations
\cite{glove:2014} are trained on aggregated global word-word co-occurrence
statistics from a corpus.
%, and the resulting representations show
%interesting \nascomment{this sounds like BS ... interesting according to who?}
%linear substructures of the word vector space.
These vectors were trained on 6
billion words from Wikipedia and English Gigaword and are of length
300.\footnote{\url{http://www-nlp.stanford.edu/projects/glove/}}

\paragraph{Skip-Gram (SG).}
The word2vec tool~\cite{mikolov2013efficient} is fast and
widely-used. In this model, each word's Huffman code is used as an
input to a log-linear classifier with a continuous projection layer and words
within a given context window are predicted. These vectors were trained
on 100 billion words %\nascomment{ungrammatical:}
of Google news data and are of length
300.\footnote{\url{https://code.google.com/p/word2vec}}

\paragraph{Global Context (GC).}
These vectors are learned using a recursive neural network that
incorporates both local and global (document-level) context features
\cite{huang2012improving}. These vectors were trained on the first 1 billion
words of English Wikipedia and are of length
50.\footnote{\url{http://nlp.stanford.edu/~socherr/
ACL2012_wordVectorsTextFile.zip}}

\paragraph{Multilingual (Multi).}
\newcite{faruqui-dyer:2014:EACL2014} learned vectors by first
performing SVD on text in different languages, then applying canonical
correlation analysis on pairs of vectors for words that align in
parallel corpora. These vectors were trained on WMT-2011 news corpus
containing 360 million words and are of length
48.\footnote{\url{http://cs.cmu.edu/~mfaruqui/soft.html}}

\section{Evaluation Benchmarks}
\label{sec:benchmarks}

Our comparisons of word vector quality consider five benchmark tasks. We now
describe the different evaluation benchmarks for word vectors.

\paragraph{Word Similarity.}
We evaluate our word representations on two word similarity tasks.
 The first is the WS-353 dataset~\cite{citeulike:379845},
which contains 353 pairs of English words that have
been assigned similarity ratings by humans. This dataset is used to tune sparse
vector learning hyperparameters (\S\ref{sec:tune}), while the remaining of the
tasks discussed in this section are completely held out.

A more recent dataset, SimLex-999 \cite{HillRK14}, has been
constructed
to specifically focus on similarity (rather than relatedness).
It contains a balanced set of noun, verb, and adjective pairs.
We calculate cosine similarity between the vectors of two words
forming a test item and report Spearman's rank correlation coefficient
\cite{citeulike:8703893} between the rankings produced by our model against the
human rankings.

\paragraph{Sentiment Analysis (Senti).}
\newcite{Socher-etal:2013} created a treebank of sentences
annotated with fine-grained sentiment labels on phrases and sentences
from movie review excerpts.
The coarse-grained treebank of positive and negative
classes has been split into training, development, and test datasets
containing 6,920, 872, and 1,821 sentences, respectively.
We use average of the word vectors of a given sentence as feature for
classification. The classifier is tuned on the dev.~set and accuracy is reported on
the test set.

\paragraph{Question Classification (TREC).}
As an aid to question answering, a question may
be classified as belonging to one of many question
types. The TREC questions dataset involves six
different question types, e.g., whether the question
is about a location, about a person, or about some
numeric information \cite{Li:2002}. The
training dataset consists of 5,452 labeled questions, and
the test dataset consists of 500 questions.
An average of the word vectors of the input question is used as features
and accuracy is reported on the test set. %\nascomment{dev set for tuning?}
%\mfar{This task has no dev test, everyone tunes and reports on test.}

\paragraph{20 Newsgroup Dataset.}
We consider three binary categorization tasks from the 20 Newsgroups
dataset.\footnote{\url{http://qwone.com/~jason/20Newsgroups}}
Each task involves categorizing a
document according to two related categories with
training/dev./test split in accordance with \newcite{yogatama2014linguistic}:
 (1) Sports:  baseball vs.~hockey (958/239/796)
(2) Comp.: IBM vs.~Mac (929/239/777) (3) Religion: atheism vs.~christian
(870/209/717).
We use average of the word vectors of a given sentence as features.
The classifier is tuned on the dev.~set and accuracy is reported on
the test set.

\paragraph{NP bracketing (NP).} \newcite{lazaridou-vecchi-baroni:2013:EMNLP}
constructed a dataset from the Penn Treebank \cite{marcus1993building} of
noun phrases (NP) of length three words, where the first can be an adjective or
a noun and the other two are nouns. The task is to predict the correct
bracketing in the parse tree for a given noun phrase. For example,
\textit{local (phone company)} and \textit{(blood pressure) medicine} exhibit
\textit{right} and \textit{left} bracketing, respectively.
We append the word vectors
of the three words in the NP in order and use them as features for binary
classification. The dataset contains 2,227 noun phrases split into 10 folds.
%We use logistic regression classifier with $\ell_2$ regularisation.
The classifier is tuned on the first fold and cross-validation accuracy
is reported on the remaining nine folds.

\bibliographystyle{acl}
\bibliography{references}
\end{document}